\documentclass{article}
\usepackage{spconf,amsmath,graphicx}
\usepackage[export]{adjustbox}
\usepackage[table,xcdraw]{xcolor}
\usepackage{adjustbox}
\usepackage{multirow}
\usepackage{float}
\title{An error correction scheme for improved air-tissue boundary in real-time MRI video for speech production}
\name{Anwesha Roy, Varun Belagali, Prasanta Kumar Ghosh}
\address{Electrical Engineering, Indian Institute of Science (IISc), Bangalore-560012, India}
\begin{document}
\ninept
\maketitle
\begin{abstract}
%-- 3DCNN the best so far
%-- However, evaluation is done using entire contour dtw
%-- which is not a right metric
%-- careful examination reveal blah blah error which are not captured in such metric
%-- in this work we automatically detect such errors and propose a correction scheme for the same
%-- we also propose new evaluation metric to quantify the quality of predicted ATB

The best performance in Air-tissue boundary (ATB) segmentation of real-time Magnetic Resonance Imaging (rtMRI) videos in speech production is known to be achieved by a 3-dimensional convolutional neural network (3D-CNN) model. However, the evaluation of this model, as well as other ATB segmentation techniques reported in the literature, is done using Dynamic Time Warping (DTW) distance between the entire original and predicted contours. Such an evaluation measure may not capture local errors in the predicted contour. Careful analysis of predicted contours reveals errors in regions like the velum part of contour1 (ATB comprising of upper lip, hard palate, and velum) and tongue base section of contour2 (ATB covering jawline, lower lip, tongue base, and epiglottis), which are not captured in a global evaluation metric like DTW distance. In this work, we automatically detect such errors and propose a correction scheme for the same. We also propose two new evaluation metrics for ATB segmentation separately in contour1 and contour2 to explicitly capture two types of errors in these contours. The proposed detection and correction strategies result in an improvement of these two evaluation metrics by 61.8\% and 61.4\% for contour1 and by 67.8\% and 28.4\% for contour2. Traditional DTW distance, on the other hand, improves by 44.6\% for contour1 and 4.0\% for contour2.
\end{abstract}
\vspace{-1.5mm}
\vspace{-1mm}
\section{Introduction}
\label{sec:intro}
\vspace{-0.6mm}

Real-time Magnetic Resonance Imaging (rtMRI) is a tool
used exhaustively in speech science and linguistics to understand the speech production process in depth across languages and health conditions \cite{hagedorn2019engineering}. rtMRI has two advantages over other methods like X-ray \cite{wold1985generation}, Electromagnetic articulography \cite{maurer1993re} and Ultrasound \cite{watkin1989pseudo} - it is safe and non-invasive, and it captures a complete picture of the vocal tract including pharyngeal structures \cite{bresch2008seeing}. 
A common step before using these rtMRI videos is obtaining the Air-Tissue Boundary (ATB) segmentation in every frame. Many works have used ATBs for different applications like text-to-speech synthesis \cite{toutios2016articulatory}, speaker verification \cite{prasad2015estimation}, visual augmentation for synthesized articulatory videos \cite{chandana2018automatic}, and analysis of vocal tract movement \cite{ramanarayanan2013investigation,lammert2013interspeaker}. The accurate estimation of ATBs of the upper airway of the vocal tract is also essential for many other speech processing applications \cite{parrell2014interaction,hsieh2013pharyngeal}. Hence, it is necessary to have an accurate and proper ATB segmentation in every frame of the rtMRI videos.

Many works in the past have addressed the problem of ATB segmentation in rtMRI frames using several supervised \cite{ mannem2019segnet,mannem2020air} and unsupervised approaches \cite{kim2014enhanced}. Although the supervised algorithms have been shown to provide accurate ATBs in the seen subject condition, a major challenge is that there is high variability in the morphology of different subjects and it is difficult to generalize ATBs in unseen subject conditions. The best performance in ATB segmentation is observed in the work by Renuka et al. \cite{mannem2020air}, where the authors tackle this problem by using a 3-dimensional deep convolutional neural network (3D-CNN) for ATB segmentation. Although the temporal continuity criterion of 3D-CNN ensures that the ATBs do not vary drastically in successive frames of a rtMRI video, in the event that there is an error in one frame, it also propagates to surrounding frames.
\vspace{-2mm}
\begin{figure}[htb]

\begin{minipage}[b]{.48\linewidth}
  \centering
  \centerline{\includegraphics[width=2.8cm]{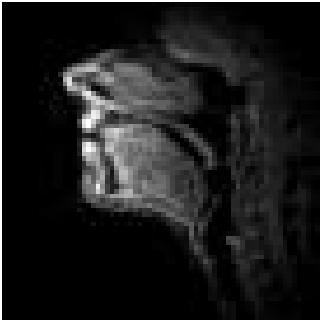}}
  \centerline{(a)}
  %\adjincludegraphics[height=5cm,trim={0 0 {.5\width} 0},clip]{mri1.eps}
%  \vspace{2.0cm}
  %\centerline{(a) Manually annotated C1}\medskip
\end{minipage}
\hfill
\begin{minipage}[b]{.48\linewidth}
  \centering
  \centerline{\includegraphics[width=2.8cm]{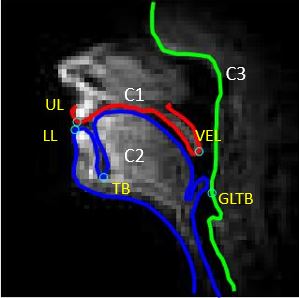}}
  \centerline{(b)}
%  \vspace{2.0cm}
 % \centerline{(b) Error in predicted C1}\medskip
\end{minipage}
\vspace{-1.3mm}
\caption{Illustration of (a) a rtMRI frame, (b) manually annotated Air Tissue Boundaries in it}
\label{fig:manual}
\end{figure}
\vspace{-5.1mm}

\begin{figure}[htb]

\begin{minipage}[b]{1\linewidth}
  \centering
  \centerline{\includegraphics[width=8.5cm]{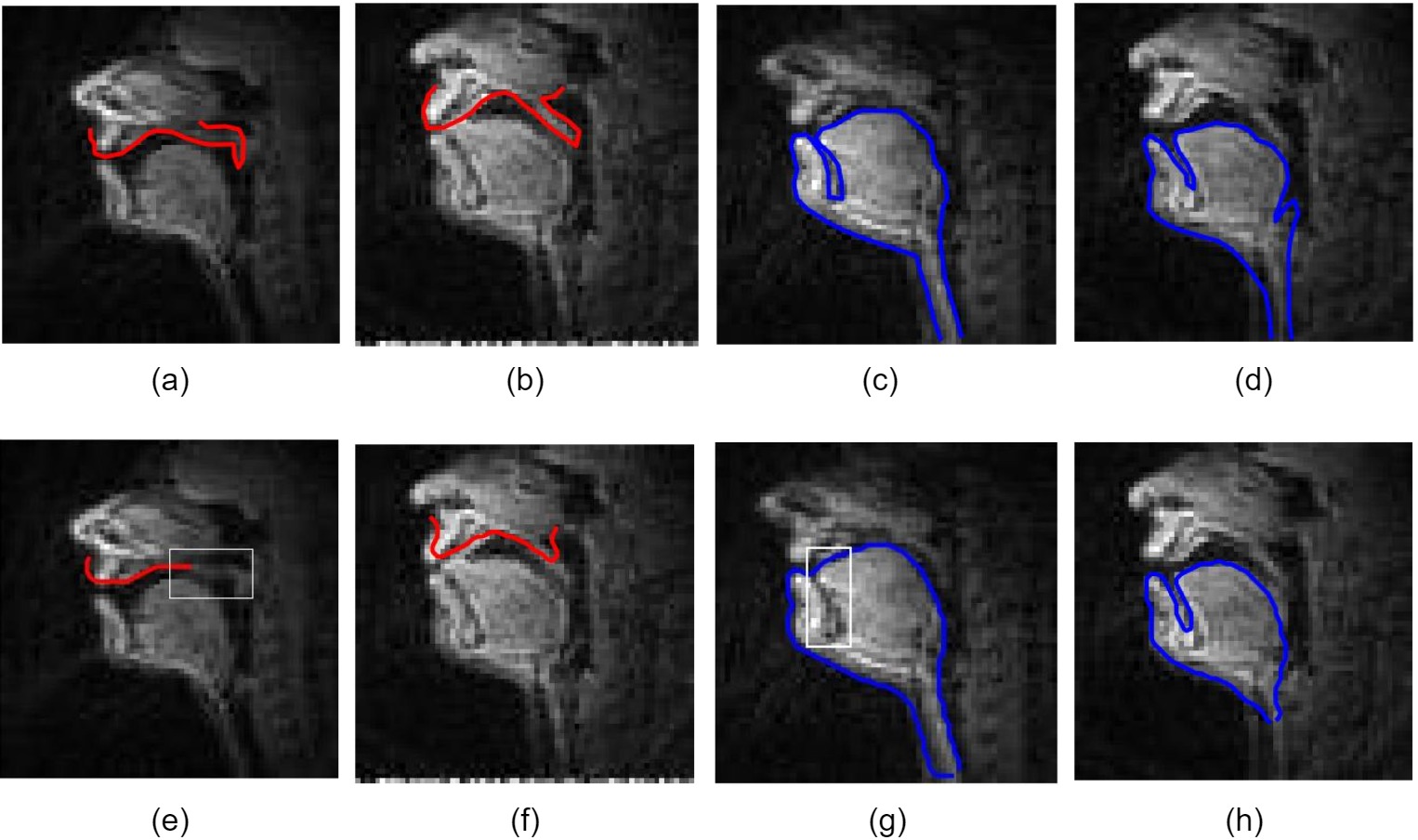}}
%  \vspace{2.0cm}
 % \centerline{Bubble plot of TB distance between annotation and prediction}%\medskip
\end{minipage}
\vspace{-5.3mm}
\caption{Manual annotations (a,b,c,d) and corresponding erroneous predictions for C1 incomplete (e), C1 frame (f), C2 TB (g) and C2 frame (h) errors}
\label{fig:error_types}
\end{figure}

\begin{figure*}[htb]

%\begin{minipage}[b]{1\linewidth}
\begin{minipage}[]{1\linewidth}
  \centering
  \centerline{\includegraphics[width=0.95\textwidth]{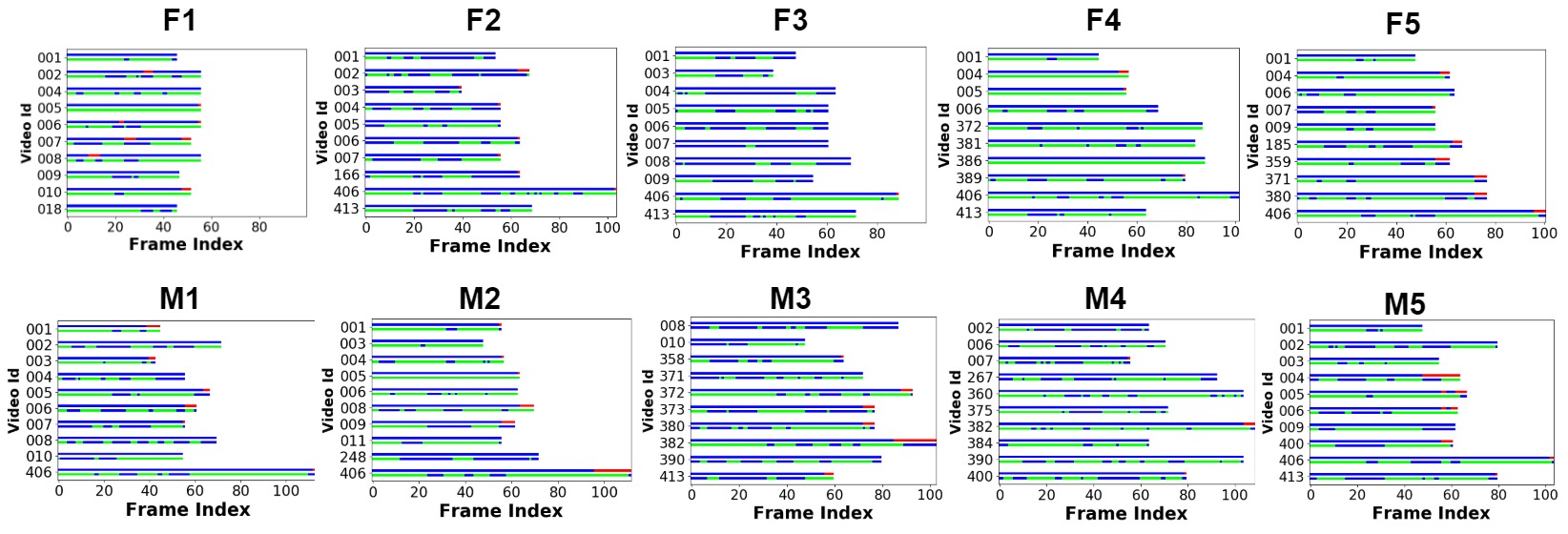}}
 %\centerline{Bubble plot of TB distance between annotation and prediction}%\medskip
\end{minipage}
\vspace{-1.3mm}
\caption{Illustration of position of error frames in each video for all subjects. There are two line plots for each video index, where each line shows correct frames in blue. C1 errors are shown in red in first line and C2 errors in green in second line.}
\label{fig:lineplot}
\end{figure*}
\vspace{-1mm}

Figure \ref{fig:manual} illustrates the three manually annotated contours, contour1 (C1), contour2 (C2) and contour3 (C3) in an rtMRI frame. We have observed that errors in contour1 can be categorized into two classes as shown in Figure \ref{fig:error_types} (e and f) - incomplete contours (where velum (VEL) portion is missing) and frame errors where the entire C1 has defects. For contour2, we again observe two types of errors: error in the tongue base (TB) region which is illustrated in Figure \ref{fig:error_types} (g), where TB dip is not predicted properly and frame error, where entire frame is wrongly predicted (Figure \ref{fig:error_types} (h)). The TB dip errors occur mainly because of the low number of pixels with low intensity present in the dip region.

In cases like these, a global measure like overall DTW distance between annotation and prediction, which has been used as an evaluation metric in previous works, does not reflect which region has what kind of error. So it is essential to conduct a region specific analysis and define evaluation metrics that focus on defects in particular regions like VEL or TB.

We have analyzed the different types of errors in the ATBs in a set of 100 videos in a subjective manner. We have proposed methods to detect and correct each of these errors. Detection methods for C1 and C2 are based on positions of VEL and TB points respectively; and correction methods focus on improving the quality of contour in the region surrounding these points. Post correction, we have developed region specific metrics for the evaluation of the quality of ATB. In order to evaluate the accuracy of the predicted C2 in the region around TB, the Euclidean distance (denoted by ETB) between annotated and predicted TBs, and regional DTW distance in the TB dip area (TBrDTW) are proposed as metrics. For C1, as the problem most often lies in the VEL region, we propose Euclidean distance (denoted by EVEL) between model predicted and ground truth VEL points, and regional DTW distance in the VEL section (VELrDTW) as the evaluation metrics. It is seen that using regional metrics helps in identifying specific problems in frames which might not be possible using global measures like DTW distance which evaluate the entire contour as a whole. For C1, after correction, the average EVEL reduces from 8.10 to 3.09 pixels, and the mean VELrDTW across frames goes down from 4.12 to 1.59 pixels. Overall DTW distance also decreases from 2.04 to 1.13 pixels. For the case of C2, overall DTW distance goes down slightly (2.05 to 1.98 pixels) post-correction, but ETB reduces substantially from 11.31 to 3.64 pixels and TBrDTW from 4.26 to 3.05 pixels.

\vspace{-2mm}

\section{Dataset}
\vspace{-2mm}
\label{sec:format}
USC-TIMIT corpus \cite{narayanan2014real} consisting of rtMRI videos of the upper airway in the mid-sagittal plane is used in this work. There are 5 female (F1, F2, F3, F4, F5) and 5 male (M1, M2, M3, M4, M5) subjects, each of them speaking 460 sentences from MOCHA-TIMIT database \cite{wrench2000multichannel}. The videos are recorded at a frame rate of 23.18 frames/sec.  Each rtMRI frame has a spatial resolution of 68 $\times$ 68 (pixel dimension of 2.9mm $\times$ 2.9mm).

For this work, 3D-CNN is trained on 90 videos (9 videos from each of the 10 subjects) using the model described in \cite{mannem2020air} and the ATBs are predicted on a set of 100 videos (10 videos from each subject) not seen in training. There are 33 unique sentences in this randomly chosen list of 100 videos. For these 100 videos (6738 frames), the ATBs are manually annotated using a MATLAB Graphical User Interface \cite{pattem2018optimal}. The manual annotation is done for three ATBs - contour1 (C1), contour2 (C2), and contour3 (C3), as well as five points that indicate upper lip (UL), lower lip (LL), tongue base (TB), velum (VEL) and glottis begin (GLTB) as illustrated in Figure \ref{fig:manual}. C1 is a closed contour that starts from UL, goes through the hard palate till VEL and goes around the fixed nasal tract. C2 is a closed contour that covers the jawline, LL, tongue blade and extends below the epiglottis. The C3 contour marks the pharyngeal wall.

%\vspace{-2mm}
\section{Error Analysis}
%\vspace{-2mm}
A graphical user interface using MATLAB is developed to observe both annotation and prediction in each frame. If prediction deviates a lot in any region from the annotation, the frame is declared erroneous and the contour where error is observed is noted (C1 or C2). Predicted C3 is not found to have any observable defects. The observations are then cross-checked by an unbiased viewer. Erroneous frames selected based on subjective criteria are considered as ground truth for error classification.

In Figure \ref{fig:lineplot}, for each video for all subjects, two line-plots are constructed. The vertical axis shows the video (sentence) ID and horizontal axis shows frame index. The first line plot illustrates correct frames in blue and C1 errors in red. The second line plot shows correct in blue and C2 error frames in green. It is observed that C2 (especially near TB) errors occur repeatedly throughout the video. C1 errors, on the other hand, are mostly observed at the end of videos. This may be due to the end frame padding done in 3D-CNN model during prediction.
\begin{table*}[!htb]
%\begin{table*}[h]
\begin{minipage}[b]{1.0\linewidth}
\caption{Summary of proposed metric, detection and correction schemes}
%\begin{LARGE}  
\label{table:summary}
\centering
%\vspace{2mm}
\setlength{\extrarowheight}{2pt}
\resizebox{\textwidth}{!}{%
\begin{tabular}{|c|c|c|c|c|}
\hline
                                   & \textbf{Error type} & \textbf{Evaluation metric}           & \textbf{Detection method}                                                                                           & \textbf{Correction method}            \\ \hline
\multirow{2}{*}{\textbf{Contour1}} & Incomplete          & \multirow{2}{*}{EVEL,  VELrDTW} & \multirow{2}{*}{\begin{tabular}[c]{@{}c@{}}Deviation from mean VEL,\\ VEL to pharyngeal wall distance\end{tabular}} & Interpolation  +  Appending           \\ \cline{2-2} \cline{5-5} 
                                   & Frame               &                                      &                                                                                                                     & Interpolation                         \\ \hline
\multirow{2}{*}{\textbf{Contour2}} & TB                  & \multirow{2}{*}{ETB,  TBrDTW}  & LL to TB slope,  LL to TB distance,  Combined                                                                       & Otsu thresholding  +  Contour warping \\ \cline{2-2} \cline{4-5} 
                                   & Frame               &                                      & No. of points                                                                                                       & Interpolation                         \\ \hline
\end{tabular}}
%\end{LARGE}
\end{minipage}
\end{table*}

%\vspace{-4mm}

%\vspace{-7mm}
\subsection{Contour 1}
\label{ssec:c1 error}
In a frame, if the VEL part of the contour is not predicted properly or there is noticeable deviation of entire contour from annotated contour or C1 is incomplete, then the frame is declared to have C1 error. As per the subjective analysis illustrated in Figure \ref{fig:lineplot}, 207 frames (3.07 \%) have C1 error. In both types of C1 error illustrated in Figure \ref{fig:error_types}, the VEL point is found to be incorrect. 
%\vspace{-3mm}

%Analysis of these frames also shows that DTW distance between annotated C1 and predicted C1 is greater for most of the frames declared erroneous subjectively (2.15 (1.48)) compared to frames declared correct (1.11 (0.18)) for all subjects(Figure \ref{fig:c1_analysis} (a)). But DTW distance of some erroneous frames is low

\begin{figure}[H]
\begin{minipage}[b]{0.47\linewidth}
  \centering
  \centerline{\includegraphics[width=4.0cm]{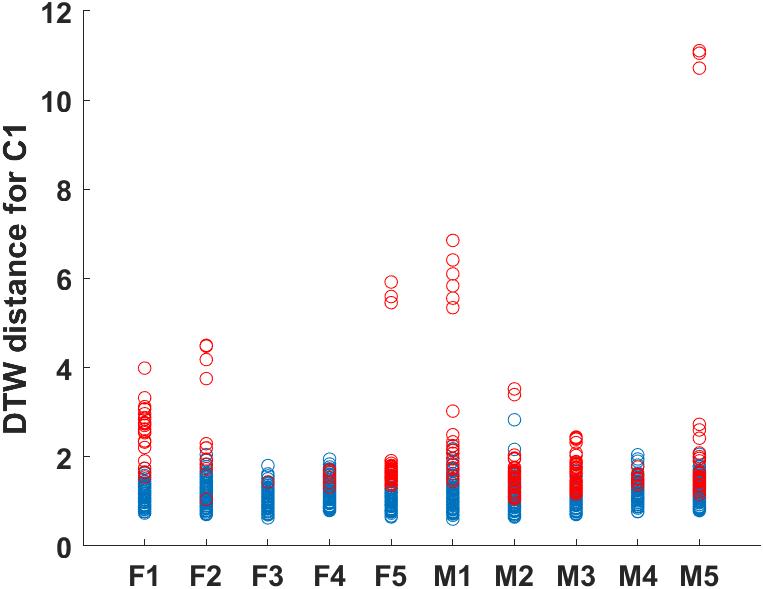}}
 % \vspace{1.5cm}
  \centerline{(a)}\medskip
\end{minipage}
\hfill
\begin{minipage}[b]{0.47\linewidth}
  \centering
  \centerline{\includegraphics[width=4.0cm]{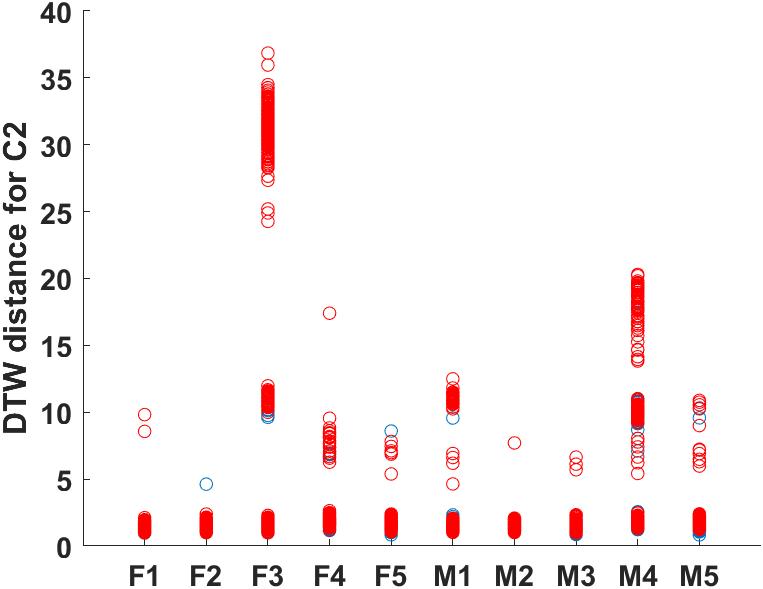}}
 % \vspace{1.5cm}
  \centerline{(b)}\medskip
\end{minipage}
\vspace{-5mm}
\caption{DTW distance analysis plots for C1 (a) and C2 (b) over all frames, where red bubbles represent erroneous frames and blue are correct ones}
\label{fig:c1_analysis}

\end{figure}

Analysis of these frames also shows that over all subjects, mean $\pm$ standard deviation (std) DTW distance between annotated C1 and predicted C1 for the frames declared erroneous subjectively is 2.15 $\pm$ 1.48 pixels compared to 1.11 $\pm$ 0.18 pixels for frames declared correct (Figure \ref{fig:c1_analysis} (a)). But the range of DTW distance of the erroneous and correct frames overlap.
\vspace{-1.5mm}

%\begin{figure}[htb]

\vspace{-7mm}
\subsection{Contour 2}
\vspace{-2mm}
The subjective analysis shows that in about 325 frames (4.82 \%), the predicted contour2 has shapes very different from expected shape (frame error) or has very low number of points. Moreover, in about 65.73 \% of the frames, the TB dip region is not predicted properly.

The DTW distance analysis shown in Figure \ref{fig:c1_analysis} (b) for C2 illustrates that erroneous and correct frames have highly overlapping range of DTW distance values. Mean $\pm$ std DTW distance for erroneous frames is 3.61 $\pm$ 3.49 pixels and for correct frames is 1.92 $\pm$ 1.89 pixels. There is high variability in the global DTW distance value and it does not reflect C2, especially TB dip, errors.

%We find the TB in the predicted C2 by locating the dip in the region between LL and tongue. For TB errors, it is observed that the Euclidean distance between TB points in annotation and prediction is substantially higher in subjectively erroneous frames than other frames (shown in Figure \ref{fig:c2_analysis}). This shows that TB distance is a good metric for C2 TB errors. It is also seen that the distance between LL to TB differs between annotation and prediction by 4.55 on average for non-erroneous frames and by 9.19 for TB error frames. 
%For frame errors, it is observed that the number of points in the predicted contour is low in itself.

%\vspace{-4mm}

%\label{sec:pagestyle}

%\vspace{-2mm}

\section{Improved ATB using proposed error correction}
\vspace{-2mm}
The proposed evaluation metric, error detection and correction methods undertaken in this work are summarized in Table \ref{table:summary}. The methods are explained in detail in the following subsections.

\vspace{-3mm}
\subsection{Experimental Setup}\label{experimental setup}
The experiments are carried out in a 4 fold cross-validation set-up where, in each fold, 3 randomly chosen subjects are taken for validation and the rest of the 7 subjects for test. The results are reported on the test set. 

We predict the VEL point by finding the dip in the row index of the contour in the region around velum in C1 and also find the TB in C2 by locating the dip in the region between LL and uppermost point on tongue.

\vspace{-4mm}
\subsubsection{\textbf{Evaluation metric}}
\vspace{-1mm}
F-score \cite{sokolova2006beyond} is used as an evaluation metric for error detection methods because of data imbalance in the set of videos considered, where only 3.07\% frames have C1 errors. For error correction of C1, VELrDTW distance is computed around the VEL region, taking 30\% of total number of points of C1 from the pharyngeal wall end. TBrDTW for C2 is found by calculating the DTW distance between annotated and predicted C2 in the region between LL and uppermost point on the tongue. Further, for C1, the Euclidean distance between predicted and ground truth VEL position (EVEL) is used for assessment. Similarly for C2, the Euclidean distance between the predicted and ground truth TB position (ETB) is also used as an evaluation metric. These metrics capture region specific errors observed during analysis. Additionally, global DTW is reported for both contours.
%For error correction, regional DTW (rDTW) distance is used as a metric for evaluating the quality of contours for both C1 and C2. rDTW distance is computed around the VEL region for C1 (VELrDTW) and tongue base region for C2 (TBrDTW). Further, for C1, the Euclidean distance between predicted and ground truth VEL position (EVEL) is also used for assessment. Similarly for C2, the Euclidean distance between the predicted and ground truth TB position (ETB) is also used as an evaluation metric. Additionally, global DTW is also reported for both contours.
\vspace{-3.5mm}
\subsection{Error Detection}
\vspace{-1.5mm}
\label{sec:detection}
We propose different methods of error detection for C1 based on relative position of VEL point. Similarly, error detection is done for C2 on the basis of position of TB and number of points. The hyperparameters used by the methods in each fold are selected based on F-score achieved on validation set.
%The validation set for each fold is used to find the hyperparameters used by the methods, as described in the upcoming sections. 

\vspace{-3mm}
\subsubsection{\textbf{Contour 1}}
\vspace{-1mm}
%\paragraph{Deviation from mean velum}
4.2.1.1  Deviation from mean velum
\vspace{1mm}
\hfill \break
For each video, Euclidean distance between velum point in a frame and mean velum point across the video is used as a measure to detect erroneous frame. This method is based on the observation in Figure \ref{fig:lineplot} that the C1 errors occur at the end of videos in most cases and are limited in number in a video. The variation in velum point coordinates across the rest of the video is low. Any frame with a significantly large distance between velum point and the mean velum point is likely to have a C1 error. A threshold is applied on this distance to classify the frame as erroneous. The thresholds selected based on  validation data are 3.5, 4, 4, 4 pixels for four folds. F-score of 0.86 ($\pm$ 0.01) is achieved on the test set.  
\vspace{2mm}
%\paragraph{Distance of velum from pharyngeal wall}
\hfill \break
4.2.1.2  Deviation of velum from pharyngeal wall
\vspace{1mm}
\hfill \break
If a video has many C1 errors then the mean velum can be affected by a cluster of erroneous velum points and can itself be wrong.  Hence, instead of using mean velum point as reference, in this method we use the nearest point on C3 from the velum. This nearest point is found out and fixed for each subject from the manual annotations used while training the 3D-CNN model. We threshold the distance between this nearest point on C3 and predicted VEL to detect erroneous frames. The thresholds selected based  on validation data are 8, 8, 8, 7.5 pixels for four folds. This method achieves an F-score of 0.85 ($\pm$ 0.02) on the test set.
%\vspace{-3mm}
%\paragraph{Combined}
\vspace{2mm}
\hfill \break
4.2.1.3  Combined
\vspace{1mm}
\hfill \break
In this method a frame is declared erroneous if it satisfies either one of the two aforementioned error criteria. The combined method achieves an F-score of 0.86 ($\pm$ 0.02) on the test set.

\vspace{-2mm}
\subsubsection{\textbf{Contour 2}}

%\paragraph{Number of points}
%\vspace{2mm}
%\paragraph{Distance of velum from pharyngeal wall}
%\hfill \break
4.2.2.1  Number of points
\vspace{1mm}
\hfill \break
For the small number of frames (4.82\%) where the predictions are very poor, it is observed that number of points in the predicted C2 is low. So, we find the average number of points in C2 over all videos in validation set and threshold the number of points at 65\% of it to find erroneous frames. This method of thresholding gives a detection F-score of 0.941 ($\pm$ 0.02) on test set.
%\vspace{-3mm}
%\paragraph{LL to TB slope}
\vspace{2mm}
\hfill \break
4.2.2.2  LL to TB slope
\vspace{1mm}
\hfill \break
TB error occurs when the groove between the lower lip and tongue is not predicted in C2. In such cases, the slope of the line joining TB and lower lip is observed to be low. We threshold this slope to detect the erroneous frames. The thresholds (0.7, 0.7, 0.8, 1) are selected based on validation data for each of the four folds. Using this method, an F-score of 0.85 ($\pm$ 0.02) is achieved on the test set.
%\vspace{-3mm}
%\paragraph{LL to TB distance}
\vspace{2mm}
\hfill \break
4.2.2.3  LL to TB distance
\vspace{1mm}
\hfill \break
When a TB error is observed, it is seen that the distance from the lower lip to TB is short. A threshold on this distance is selected based on F-score achieved on validation data and any frame with distance higher than it are declared erroneous. The thresholds selected are 8, 7, 10, and 10 pixels for four folds. The F-score on the test set is 0.88 ($\pm$ 0.02) for this method. The distance thresholding performs better than slope.
%\vspace{-4mm}
%\paragraph{Combined}%\label{TB detection}
\vspace{2mm}
\hfill \break
4.2.2.4  Combined
\vspace{1mm}
\hfill \break
In this method a frame is declared erroneous if it satisfies either one of the three aforementioned error criteria. F-score of 0.90 ($\pm$ 0.02) is achieved on the test set.

\iffalse
\begin{table}[!htb]
\begin{minipage}[b]{1.0\linewidth}
\caption{F-scores for detection of C1 and C2 TB errors}
\label{table:detection}
\centering
%\vspace{2mm}
\setlength{\extrarowheight}{2pt}
\resizebox{\textwidth}{!}{%
\begin{tabular}{|c|c|c|c|c|c|}
\hline
\cellcolor[HTML]{FFFFFF}{\color[HTML]{333333} } & \multicolumn{2}{c|}{C1} & \multicolumn{3}{c|}{C2 TB} \\ \hline
Fold                                            & Mean VEL        & VEL to C3       & Slope   & Distance   & Combined   \\ \hline
1                                               & 0.86            & 0.82            & 0.82    & 0.89       & 0.90       \\ \hline
2                                               & 0.85            & 0.85            & 0.87    & 0.86       & 0.89       \\ \hline
3                                               & 0.85            & 0.85            & 0.86    & 0.91       & 0.92       \\ \hline
4                                               & 0.87            & 0.87            & 0.84    & 0.86       & 0.88       \\ \hline
\end{tabular}}
%\end{adjustbox}
\end{minipage}
\end{table}
\fi
\vspace{-2mm}
\subsection{Error Correction}
\label{sec:detection}
The error correction methods proposed for contour 1 and contour 2 are detailed in the following subsection.
\vspace{-2mm}
\subsubsection{\textbf{Contour 1}}
%\vspace{-1mm}
For all erroneous frames detected using the combined method mentioned in section 4.2.1.3, C1 is generated by linear interpolation using neighbouring frame's contours as the frame to frame variance is low because of temporal continuity.
In case of the incomplete C1 errors, it is observed through section wise DTW distance analysis that the part of the predicted contour is actually correct and the rest is missing. So, for them, we find the end point of original contour on interpolated one and append the rest of the interpolated contour to the existing contour so that the upper lip part of existing C1 is not affected by interpolation. For the frame errors, the interpolated contour is taken completely.

The EVEL decreases by 61.8\% after correction as shown in Table \ref{table:c2 correction}, whereas VELrDTW improves by 61.4\% for these frames. We also observe that for all frames, the DTW distance over entire C1 improves by 44.6\% after correction, which is not as significant as the change in VELrDTW. An example of a frame with incomplete C1 error before and after correction is illustrated in Figure \ref{fig:output} (b) and (c) respectively.

\vspace{-2mm}
\subsubsection{\textbf{Contour 2}}
For all frame errors detected in section 4.2.2.1, we generate the entire contour by linear interpolation using neighbouring frame contours. Next, these frames along with the erroneous frames detected in section 4.2.2.4 are considered for C2 TB correction. To correct C2 near TB, we first correct the position of TB. The TB is observed to be within a 15 $\times$ 20 patch of the frame in the low intensity air cavity region between lip and tongue. We apply Ostu thresholding \cite{otsu1979threshold} to this 15 $\times$ 20 patch in the rtMRI video frame to find darker pixels that lie within C2. The result of Ostu thresholding is a binary image, where class-1 corresponds to tissue and class-0 is for the air cavity. The lowest point in the class-0 region of the binary image within C2 is selected and marked as corrected TB. Next, we adjust C2 in the vicinity of the 3D-CNN predicted TB location by warping the existing contour. We find the shift in TB point from prediction to correction and map the shift of the neighbouring points accordingly in a gradient based fashion to find corrected C2 in TB dip region.

\begin{figure}[htb]

\begin{minipage}[b]{1.0\linewidth}
  \centering
  \centerline{\includegraphics[width=8.5cm]{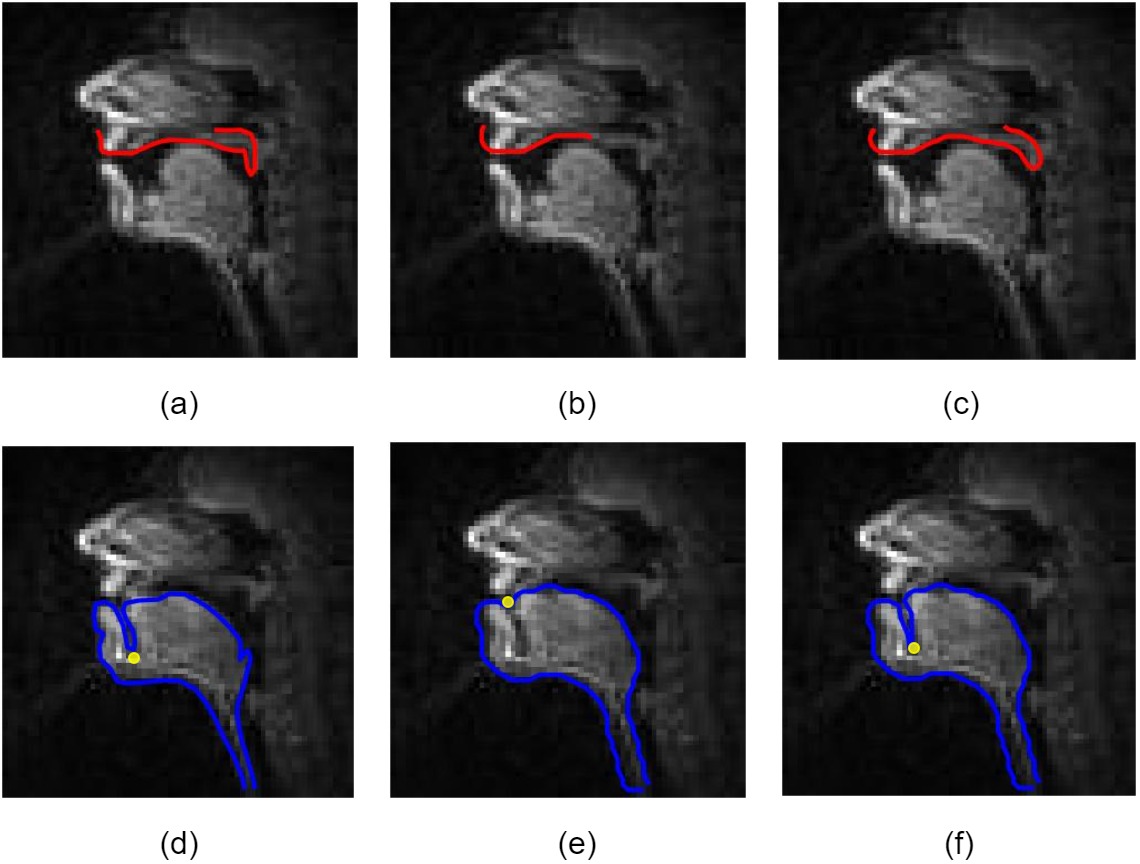}}
%  \vspace{2.0cm}
 % \centerline{Bubble plot of TB distance between annotation and prediction}%\medskip
\end{minipage}
\vspace{-4mm}
\caption{Manual annotations (a,d), erroneous predictions (b,e) and corrected contours (c,f) for VEL and TB errors respectively, where yellow point represents TB}
\label{fig:output}
\end{figure}

The evaluation metrics - ETB and TBrDTW, improve by 67.8\% and 28.4\%, respectively, after correction as shown in Table \ref{table:c2 correction}. The global DTW distance, on the other hand, does not show any significant improvement. A frame with C2 TB error is shown before and after correction in Figure \ref{fig:output} (e) and (f) respectively.

\vspace{-5mm}
\begin{table}[!htb]
\begin{minipage}[b]{1.0\linewidth}
\caption{Mean $\pm$ standard deviation of evaluation metrics (in pixels) before and after correction for C1 and C2}
\label{table:c2 correction}
\vspace{1mm}
\resizebox{\textwidth}{!}{%
\begin{tabular}{|c|c|c|c|}
\hline
\textbf{}                    & \textbf{Evaluation Metric} & \textbf{Pre-correction} & \textbf{Post-correction} \\ \hline
\multirow{3}{*}{\textbf{C1}} & EVEL                   & 8.10 $\pm$ 2.33             & 3.09 $\pm$ 1.34              \\ \cline{2-4} 
                             & VELrDTW                       & 4.12 $\pm$ 1.56             & 1.59 $\pm$ 0.43              \\ \cline{2-4} 
                             & DTW                        & 2.04 $\pm$ 1.19             & 1.13 $\pm$ 0.19              \\ \hline
\multirow{3}{*}{\textbf{C2}} & ETB                    & 11.31 $\pm$ 3.40            & 3.64 $\pm$ 2.71              \\ \cline{2-4} 
                             & TBrDTW                       & 4.26 $\pm$ 1.26             & 3.05 $\pm$ 1.06              \\ \cline{2-4} 
                             & DTW                        & 2.06 $\pm$ 1.22             & 1.98 $\pm$ 1.32              \\ \hline
\end{tabular}}
\end{minipage}
\end{table}

\vspace{-6mm}

\vspace{-6mm}
\section{Conclusion}
\vspace{-2mm}
In this work, we propose an error correction scheme for improving predicted air tissue boundaries (ATBs) in real time MRI video. Careful analysis reveals different types of errors present in the results of 3D-CNN model like VEL or TB region defects. Automatic methods are proposed to detect and correct such observed errors. Further, region specific metrics are proposed for evaluation of the quality of the predicted and corrected contours. The proposed methods show observable refinement in air tissue boundaries, which is reflected in the improvement in proposed metrics. In our future work, we will explore Active Appearance Models (AAM) \cite{cootes1998active} for correction of regional errors like TB or VEL. Further, we will try to explore robust neural network approaches using region specific loss functions, which target specific problems in particular contour regions. We will also investigate the performance of these corrected ATBs in different applications as mentioned in \cite{toutios2016articulatory, prasad2015estimation, ramanarayanan2013investigation, lammert2013interspeaker}.

%\subsection{Subheadings}
%\label{ssec:subhead}

%\subsubsection{Sub-subheadings}
%\label{sssec:subsubhead}

%\cite{Lamp86}.

% Below is an example of how to insert images. Delete the ``\vspace'' line,
% uncomment the preceding line ``\centerline...'' and replace ``imageX.ps''
% with a suitable PostScript file name.
% -------------------------------------------------------------------------

%\vfill
%\pagebreak

%\vfill\pagebreak

\bibliographystyle{IEEEtran}
%{\small
%\bibliography{refs}}

\bibliography{refs}
%\bibliography{refs}

\end{document}